\documentclass{article} 
\usepackage{iclr2023_conference,times}

\usepackage{amsmath,amsfonts,bm}

\def\eqref#1{equation~\ref{#1}}

\def\1{\bm{1}}

\DeclareMathAlphabet{\mathsfit}{\encodingdefault}{\sfdefault}{m}{sl}
\SetMathAlphabet{\mathsfit}{bold}{\encodingdefault}{\sfdefault}{bx}{n}

\usepackage{hyperref}
\usepackage{url}

\usepackage{subcaption}
\usepackage{float}

\usepackage[colorinlistoftodos]{todonotes}

\title{SynthPop++: A Hybrid Framework for Generating A Country-scale Synthetic Population}

\author{Bhavesh Neekhra, Kshitij Kapoor \& Debayan Gupta 
\\
Department of Computer Science\\
Ashoka University\\
India \\
\texttt{\{bhavesh.neekhra\_phd18,kshitij.kapoor1,debayan.gupta\}@ashoka.edu.in} \\
}
\iclrfinalcopy 

\begin{document}

\maketitle

\begin{abstract}
Population censuses are vital to public policy decision-making. They provide insight into human resource, demography, culture, and economic structure at local, regional, and national levels. However, such surveys are very expensive (especially for low and middle-income countries with high populations, such as India), time-consuming, and may also raise privacy concerns, depending upon the type of data collected.

In light of these issues, we introduce SynthPop++, a novel hybrid framework, which can combine data from multiple real-world surveys (with different, partially overlapping sets of attributes) to produce a real-scale synthetic population of humans. Critically, our population maintains family structures comprising individuals with demographic, socioeconomic, health, and geolocation attributes: this means that our ``fake'' people live in realistic locations, have realistic families, etc. Such data can be used for a variety of purposes: we explore one such use case, Agent-based modelling of infectious disease in India.

To gauge the quality of our synthetic population, we use machine learning and statistical metrics. Our experimental results show that synthetic population can realistically simulate the population for various administrative units of India, producing real-scale, detailed data at the desired level of zoom -- from cities, to districts, to states, eventually combining to form a country-scale synthetic population.
\end{abstract}

\section{Introduction}
Financial institutions, government agencies, think tanks, etc. are using techniques like agent-based modelling(ABM)~\citet{Bonabeau02} to simulate increasingly complex scenarios for decision-making. However, the collection of required datasets for model training and Agent-based modelling are constrained by real-world challenges, such as high costs of acquisition, privacy laws, etc. 

\begin{figure}[H]
\begin{center}
\includegraphics[width=0.8\linewidth]{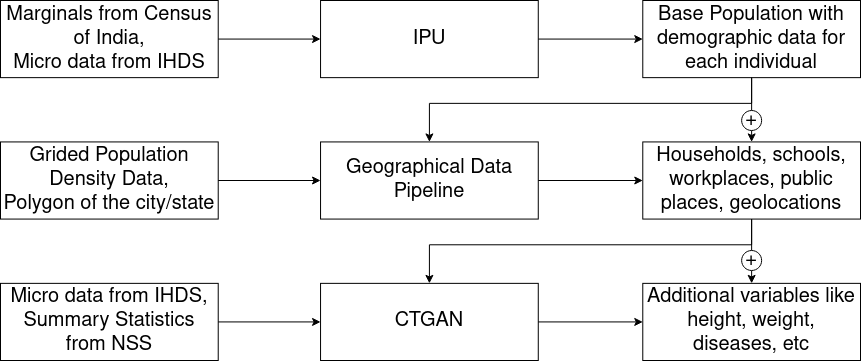} 
\end{center}
\caption{Our Framework : Pre-processed data is used as input for generating data}
\label{framework}
\end{figure}

Moreover, population censuses and various other surveys, e.g., for health, economics, education etc., help inform public policy decision-making. Periodic censuses and surveys provide aggregate statistics on variables describing the demography, socioeconomic distribution and health status of the constituents at various levels of zoom. They also help decision-makers examine the effects of previous public policy decisions on the evolution of these statistics. Surveys that provide data at an individual level instead of marginals are of particular interest as once released, they allow for superior analysis of correlations between different attributes. However, data collection for censuses and surveys is a time-consuming and expensive task, especially for low and middle-income countries. For example, the last Census in India, during 2011, involved a total of 2.7 million ``enumerators'' and ``supervisors''~\citep{census-2021}, costing approximately INR 22,000 million. Further, national censuses do not often release full-scale individual data for public access due to privacy reasons which limit the potential for analysis. 

In light of these challenges, Privacy-preserving synthetic populations can be used as an alternative to real individual-level datasets. To increase the accuracy of downstream tasks, it is essential that the marginal and joint distributions of the various attributes are similar across the original and the synthetic population. Further, to address privacy concerns, it is necessary to ensure that the synthetic dataset does not leak any personally identifiable information about the individuals who were originally surveyed.

Additionally, with the growing complexity of Agent-based models, it is necessary to generate synthetic datasets where individuals are parts of networks, and the linkage is based on shared attributes like a household, a workplace, a school, common direction of travel, etc. Preexisting frameworks for synthetic population generation are limited in this regard.

To focus our discussion, we will explore the task of generating a synthetic population with the use case of modelling disease spread (via Agent-based modelling) throughout this paper.
In this context, we make the following contributions:
\begin{itemize}
    \item A novel hybrid framework which combines various state-of-the-art statistical and machine learning models to generate a real-scale synthetic population with network and geolocation data.
    \item A combination of statistical and Machine Learning based metrics to gauge the quality of the synthetic population
    \item To promote transparency and reproducibility, our code and datasets are open-sourced\footnote{\url{https://github.com/k00lk0der1/synthpoppp/}}
\end{itemize}

The rest of the paper is organised as follows. Section 2 is about related work to generate a synthetic population. In Section 3, we describe our model in detail. The experiments and evaluation are presented in Section 4. We discuss future work and conclude in Section 5. 
\section{Related Work}
In the task of modelling and generating synthetic populations, there are various challenges in modelling the tabular data. For example, a table might have mixed data types with discrete and continuous columns. The columns in the table (especially those with continuous values) might not
follow, say, the Gaussian distribution. Further, a column might have multi-modal distributions. The data may also be highly imbalanced, resulting in insufficient training opportunities for minor classes. Some ABMs require the synthetic population to be realistically geolocated, which, to the best of our knowledge, has not been attempted, especially not for India.

Though various methods have been suggested to tackle these issues, they do not provide a solution to the issues specified. For example, Bayesian networks \citet{PGM_MIT} combine low dimensional distributions to approximate the full-dimensional distribution of a data set and are a simple but powerful example of a graphical model. But, Bayesian methods can not model tabular data effectively when both continuous and discrete columns are present. Deep Generative models are generally more sophisticated but, in practice, can not outperform Bayesian methods as the tabular data may contain Non-Gaussian continuous columns and/or imbalanced discrete columns. Some methods have tried to mitigate these issues. PrivBayes \citet{PrivBayes} can model the table with discrete variables, but all continuous variables need to be discretised. Additionally, noise injection to preserve privacy may result in low-quality synthetic data. MedGAN \citet{medgan} has been used to generate health records, but it does not support mixed data types. It also can not handle Non-Gaussian, Multi-modal, imbalanced categorical columns, lack of training data and missing data. Although CTGAN \citet{xu2019} can work with tables with mixed data types, high dimensionality, imbalanced categorical data etc., it can not handle a lack of training data and missing values. Also, it can not model realistic family structures. Iterative Proportional Updating (IPU) \citet{ipu2009} can work with limited data and model realistic family structures, but it does not scale well with high dimensional data. To tackle these issues, we propose our framework, which uses a hybrid, as shown in diagram \ref{framework} of IPU and CTGAN to take advantage of the best from these two models. Our framework allows for combining data from various sources and generates a synthetic population with required columns. 
\section{Our Framework}
Our objective is to generate a real-scale synthetic population for an entire country with distributions and joint distributions of various variables that match those of the actual population. This is necessary to ensure that the results of the downstream tasks are reliable.

A synthetic population is a limited individual-level representation of the actual population. However, not all the attributes of an individual are included (for example, hair colour or shoe size might be irrelevant for modelling disease spread, while co-morbidities like diabetes would be included). As such, a synthetic population does not aim to perfectly mimic reality -- this would be impossible. Instead, it attempts to sufficiently match various statistical measures observed in the real population.

The components of any pipeline used for generating a synthetic population would largely depend on the variables that are required for the downstream task. As stated earlier, our pipeline, as shown in diagram \ref{framework}, is built with a focus on synthetic population data for district-level infectious disease modelling.

For the chosen downstream task, the synthetic population should capture various heuristics like the size of a family;
the joint distribution of age and sex of individuals within a family; administrative units like cities, districts, and states;
geographical distribution of households, workplaces, and schools within an administrative unit;
the number of people who are associated with a workplace or school;
and various other individual-level statistical correlations that can be observed in any population.
Iterative Proportional Updating and Conditional Tabular GAN were used to generate families and individuals and their related metadata. Further, reject sampling was used to sample geo-location points within a city's boundaries which followed the per unit area population density distribution. A method for random assignment of external locations like workplaces, schools and public places was used where the probability of selecting any external location was inversely proportional to the distance between a given individual's household and the external location ~\citet{goi_2019}.
\subsection{Data Generation with Family Structure}
IPU~\citet{ipu2009} is a sampling method which ensures that the marginal distributions for both households and individuals in the synthetic population match with the marginals distributions of the real population. Hence, sampling with IPU helps generate a population in which households have a realistic distribution of the number of family members and their age and gender. However, matching only marginal distribution is not enough. We must closely resemble the joint distribution of various attributes in the population. To achieve this, we use CTGAN~\citet{xu2019} for adding individual-level attributes for which marginals are not available. The CTGAN model can be conditioned on attributes present in the base population generated by IPU to generate additional attributes. Thus using CTGAN, we can get a reasonable joint distribution of various individual-level attributes in the synthetic population. Additionally, multiple CTGAN models can also be used to join survey datasets which have an overlapping set of attributes. This allows for the expansion of attributes as per the needs of the downstream task. 

\subsection{Population Density Distribution Sampling} \label{ppds}
Given that the downstream task is to analyse the spread of infectious diseases in a given region, it is necessary to have an accurate geographical distribution of population density and workplace density. To this end, grid population density data ~\citet{gaughan2013high} is used. A row in the grid population density data consists of latitude (\(X\)), longitude (\(Y\)) and the number of individuals (\(Z\)) who live within a square with edge length \(S\) and with its centre at the given latitude and longitude (\(X\), \(Y\)). For the data set used in the current pipeline, the length of the side of square \(S=0.3\) \(arctan\) translates to a length of about \( 1\) \(km\) on the equator. The sampling method also needs a polygon for the boundaries of the region for which data is being generated. We use the geojson file for each district to get these geographical boundaries. 
\\\\
Let \(n\) be the number of latitude-longitude pairs that need to be sampled within this region. A filtered subset of the population density dataset is obtained by only retaining those rows for which the latitude \(X\) and longitude \(Y\) are within the boundary polygon of the region. We sample with replacement \(k\) latitude longitude pairs from this filtered subset and use the number of individuals(\(Z\)) as the weight for this sampling. For a given row in the sample, we then sample points within the squares corresponding to the latitude-longitude pair by adding independent random uniform noise \(A, B \sim U(-S/2, S/2)\) to the latitude \(X\) and longitude \(Y\) respectively. We reject those latitude-longitude pairs that are not within the polygon. Given this method uses reject sampling, we initially need to sample more points than required, and hence we need to use a suitable value for \(k>n\) (we choose a values \(k = 10n\)) and then choose at random from the pairs which were not rejected.
\\\\
We use this method to assign geolocation to houses, workplaces, schools and public places. Given we do not have access to workplace distribution data, we assume that it follows the population density distribution as well. \textbf{However, this assumption is not valid in cities where people live in outer sub urban areas and work closer to the city center and other possible scenarios.} With the appropriate dataset, we can substitute the population density data with workplace density data and get a distribution similar to the ground reality with the sampling method described earlier.
\subsection{External Location Assignment} \label{ela}
Since infectious disease modelling often includes analysis based on contact network graphs, individuals in this synthetic population would need to be associated with external locations like workplaces, schools and public places which they might visit periodically. This assignment is based on the assumption that adults work at workplaces, children go to schools, and people are more likely to visit public places which are closer to their homes. We use the L2 distance metric for this computation since the earth's curvature within a city is negligible, but a geodesic distance metric can also be used. Given an individual with home latitude \(X_h\) and longitude \(Y_h\) and a list of possible external locations with latitude longitude pairs (\(X_{E_{1}}\), \(Y_{E_{1}}\)), ..., (\(X_{E_{k}}\), \(Y_{E_{k}}\)), we calculate the \(L_2\) distance between the home of the individual and the external location. Then we choose a function \(f(x)\) which is strictly decreasing in the positive real domain. Then we weigh the probability of the individual being assigned the external location by the value \(f(D((X_h, Y_h), (X_{E_{k}}, Y_{E_{k}})))\) where \(D\) is the \(L_2\) distance in 2 dimensions.
\section{Experiments and Evaluation}
\subsection{Datasets}
As mentioned in \textit{Our Framework} section, we use various datasets as input to our framework. We use sample survey data (micro-data) from India Human Development Survey-II \citet{Desai18}, marginals from Census, 2011 \citet{census-2011}, Data for employment from NSS \citet{nss68}, and population density from GADM grid population density dataset \citet{j-hijmans-2018}. 
The IHDS-II, 2011-12 is a nationally representative, multi-topic survey of 42,152 households in 1,503 villages and 971 urban neighbourhoods across India. We have used micro-data from two datasets from this survey, namely \textit{Individual} and \textit{Household} as input to the CTGAN model. The Census of India 2011 was done across 28 states and 8 union territories, covering 640 districts, 5,924 sub-districts, 7,935 towns and more than 600,000 villages. We use marginal data from Census 2011 as input to IPU (The Census micro-data is not publicly available). The NSS collects data through nationwide sample household surveys on various socio-economic subjects. We have used employment data from this survey. GADM provides maps and spatial data for all countries and their sub-divisions. We used population density data for India to place the generated synthetic population across geography. In the next section, we describe how to use these different datasets as input to our hybrid model to generate a synthetic population for a district of India. 
\subsection{Generating Synthetic Population}
A synthetic population for Agent-based disease modelling should have demographic, socioeconomic, health-related and geographic 
variables for each individual in the population. We will now describe the steps for generating a synthetic population for the  
district of Mumbai, located in the state of Maharashtra in India. As discussed in the previous section, we use marginals for 
individual and household attributes from Census 2011. For individuals, the relevant attributes are age, sex \footnote{We used 
the same terminology as used in the Census of India}, religion and caste. For households, the attribute is household size, 
i.e. the number of members in the household. We also use a subset of IHDS-II dataset obtained by filtering for individuals and 
households which are situated in the state of Maharashtra. We use the filtered datasets for Maharashtra because the Mumbai 
dataset in IHDS-II contains a few hundred samples, which are not enough for our model to generate quality synthetic 
population. We then use IPU which use the marginals and the microdata to generate households of varying sizes. Each 
household has a certain number of individual family members, each with their own age, sex, religion and caste. There may be 
some shared attributes between family members like common religion and caste. Further the distribution of age and sex of 
members within a family of a given size resembles that of a typical family of the given size that one would expect to see in 
the real world. We then sample latitude longitude pairs using the method described in Section \ref{ppds} for each household. 
The same household latitude and longitude is attached to each member of the household. 
\\
We add individual level attributes like weight, height, comorbidities and job description. For weight, height and 
comorbidities, we use a CTGAN model conditioned on age and sex. The CTGAN model is trained on the subset of IHDS-
II dataset containing individuals situated within the state of Maharashtra. For individuals below the age of 3, 'Homebound' is 
assigned as the job description. For individuals above the age of 3 and below the age of 18, 'Student' is assigned as the job 
description. For individuals, the job description is sampled with replacement from the list of job descriptions of the 
individuals in the IHDS-II subset. The weight assigned to each possible job description is linearly proportional to the 
distribution observed in the IHDS-II subset for Maharashtra. 
\\
Next, we generate synthetic workplaces, schools and public places for our synthetic individuals. 
The number of workplaces, schools and public places is a parameter that varies for each district. 
Each workplace has three attributes, workplace type, latitude and longitude. The attribute 
workplace type is generated by sampling (with replacement) from the list of jobs in the IHDS-II 
subset. Latitude and longitude pairs are sampled with the method described in Section \ref{ppds}. 
Schools are obtained by filtering for a subset of workplaces with workplace types set to 
'Teacher'. Next, we assign schools to students and workplaces to individuals who are not homebound 
or students. Assigning schools in straightforward as we use the method from Section \ref{ela}.  
The final synthetic population has demographic data (e.g., age, height etc.), disease data (e.g., 
chronic heart disease, diabetes etc.), family data, geographic location for workplaces, schools 
etc., and socio-economic data(e.g., job, religion etc.). In total, we generate a synthetic population with multiple columns. A sample of our synthetic population is shown in the \textit{Appendix}. 
\subsection{Evaluating Synthetic Population}
According to Census of India, 2011 \citet{census-mumbai}, the district of Mumbai had a total population of 3,085,411, and the
district of Mumbai Suburb had a total population of 9,356,962. We generated the synthetic population for these two districts 
combined. We evaluate this synthetic population and present our results here. 

\begin{table}[H]
    \centering
    \caption{Statistical Metrics}
    \label{stats}
    \begin{tabular}{lc}
         \multicolumn{1}{c}{\bf Test}  &\multicolumn{1}{c}{\bf Score} 
            \\ \hline \\ 
            Chi-square Test for Sex             & 0.99 \\
            Kolmogorov-Smirnov for Age          & 0.98 \\
            Kolmogorov-Smirnov for Height       & 0.91 \\
            Kolmogorov-Smirnov for Weight       & 0.95 \\
    \end{tabular}
 \end{table}   
 

\begin{table}[H]
\begin{center}
\parbox{.8\linewidth}{
\caption{ML efficacy tests}
\label{ML}
\begin{tabular}{lcc}
\multicolumn{1}{c}{\bf Test}  &\multicolumn{1}{c}{\bf Source Population} &\multicolumn{1}{c}{\bf Synthetic Population}
\\ \hline \\
Linear Regression for Weight     &0.72     &0.69\\
MLP Regressor for Weight     &0.78     &0.76\\
Linear Regression for Height &0.69     &0.65\\
MLP Regressor for Height    &0.71     &0.72\\
\end{tabular}

}
\end{center}
\end{table}
Though there are no standard benchmark techniques for evaluating the quality of synthetic population, previous works have tried different methods. For example, \citet{Sync} uses ML classification tasks to check the usefulness of synthetic population augmented data. \citet{Yue2018} uses classification and regression tasks to validate the synthetic population. 
We evaluate the synthetic population based on the following three criteria:
\begin{itemize}
    \item Distribution of individual features in synthetic population match those in sample surveys, i.e., histogram should look similar
    \item Joint distribution of selected features in synthetic population match those in sample surveys, i.e., scatter plot should look similar
    \item Relationship between features and selected target variable in synthetic population match the same in sample survey
\end{itemize}
\begin{figure}[h]
\begin{center}
\includegraphics[width=1.0\linewidth]{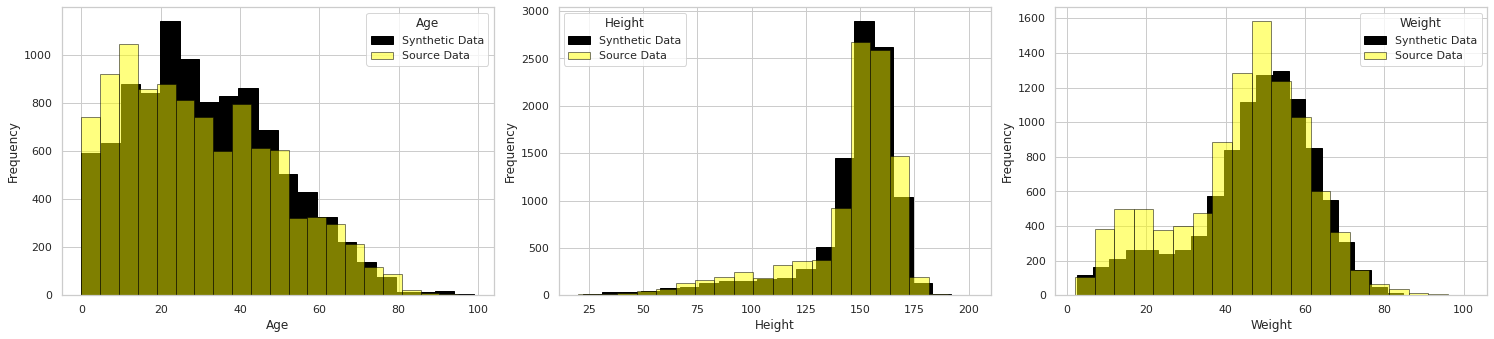} 
\end{center}
\caption{Histogram for marginal distributions of age, height and weight: Comparing source population with synthetic population for the combined districts of Mumbai and Mumbai Suburban, India}
\label{histogram_ahw}
\end{figure}

\begin{figure}[h]
\begin{center}
\includegraphics[width=1.0\linewidth]{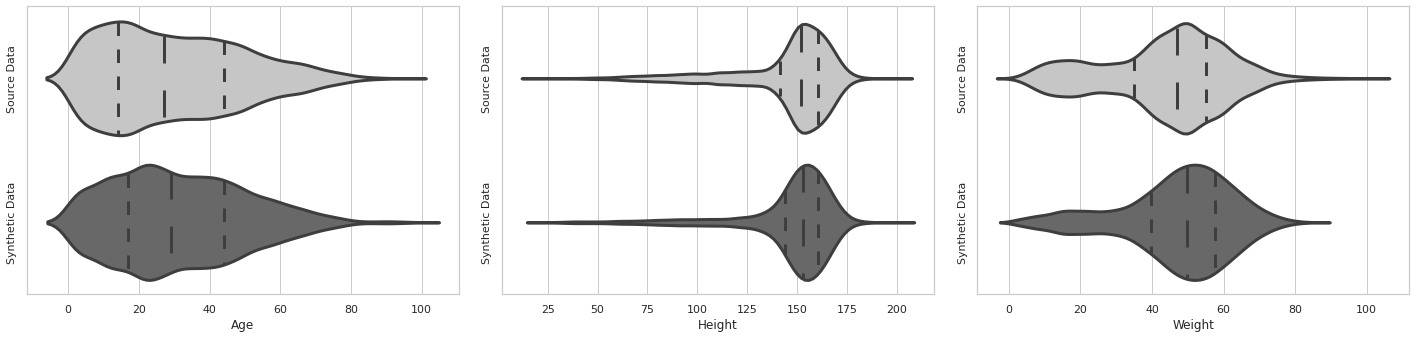} 
\end{center}
\caption{Violin plots for marginal distributions of age, height and weight: Comparing source population with synthetic population for the combined districts of Mumbai and Mumbai Suburban, India}
\label{violin_ahw}
\end{figure}

\begin{figure}[h]
\begin{center}
\includegraphics[width=1.0\linewidth]{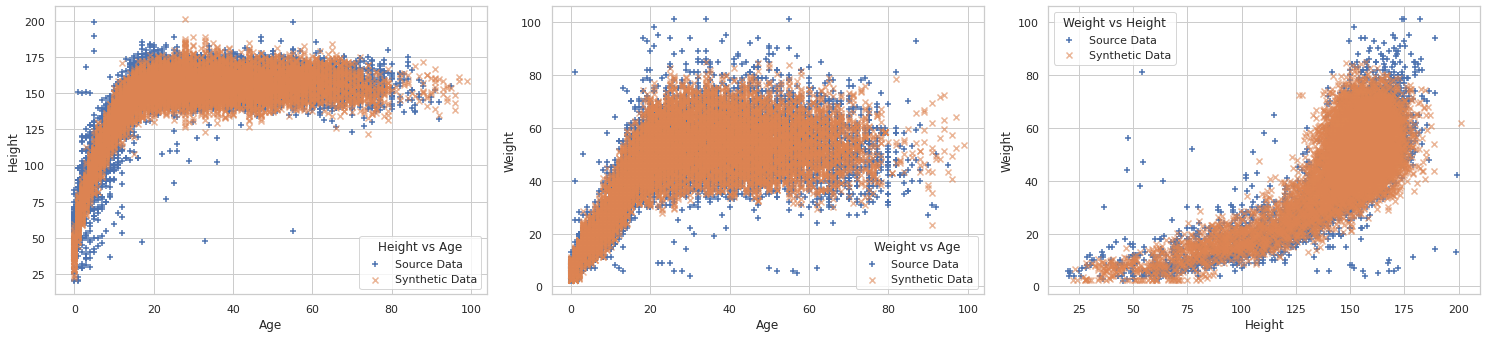} 
\end{center}
\caption{Scatter plots for joint distributions of Age, Height and Weight: Comparing source population (+) with synthetic population (x) for the combined districts of Mumbai and Mumbai Suburban, India}
\label{scatter_plot_ahw}
\end{figure}

\begin{figure}[h]
\begin{subfigure}{.33\linewidth}
  \centering
  \includegraphics[width=\linewidth]{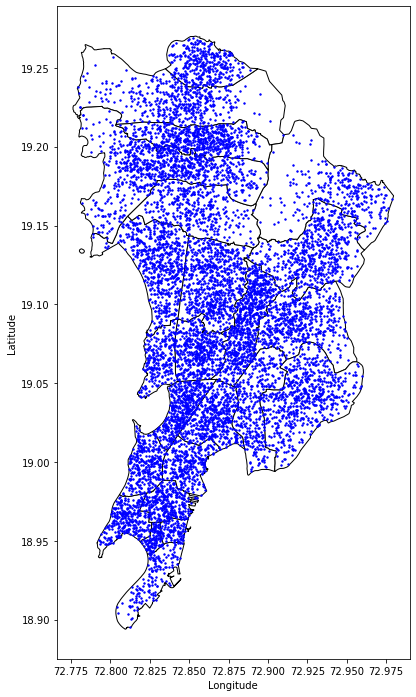}
  \caption{Households}
  \label{fig4:sfig1}
\end{subfigure}%
\begin{subfigure}{.33\linewidth}
  \centering
  \includegraphics[width=\linewidth]{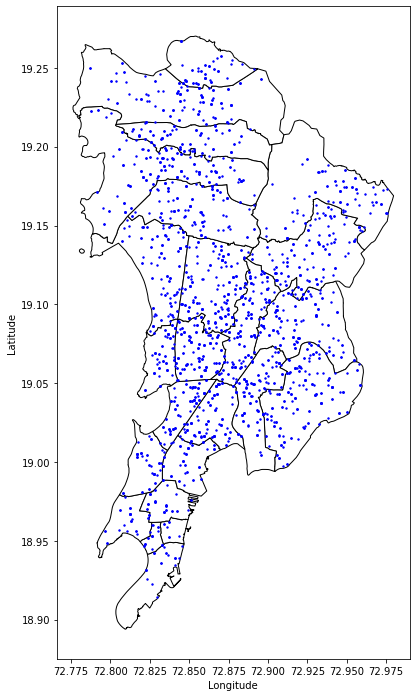}
  \caption{Schools}
  \label{fig4:sfig2}
\end{subfigure}
\begin{subfigure}{.33\linewidth}
  \centering
  \includegraphics[width=\linewidth]{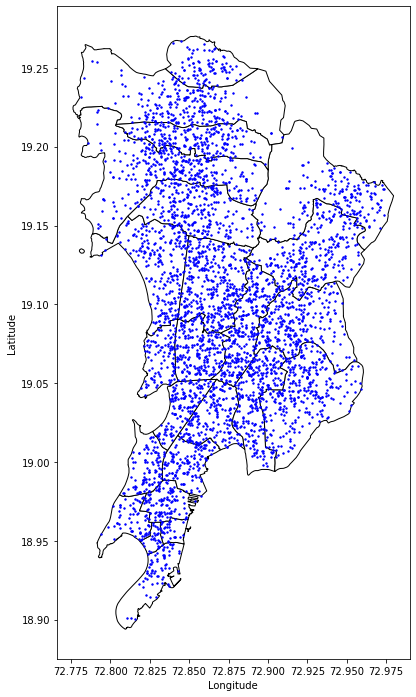}
  \caption{Workplaces}
  \label{fig4:sfig3}
\end{subfigure}
\caption{Geographical distribution of (A) Households, (B) Schools and (C) Workplaces for the combined synthetic population for the districts of Mumbai and Mumbai Suburban, India}
\label{scatter_plot_geo}
\end{figure}

Figure \ref{histogram_ahw} and \ref{violin_ahw} help compare the marginal distributions of attributes of age, height and 
weight across the real and the synthetic population for the combined districts of Mumbai and Mumbai Suburban, India. The 
marginal distribution has similar shapes even though the synthetic population is real scale. Figure \ref{scatter_plot_ahw} 
juxtaposes the scatter plots of attributes of age, height and weight for real and synthetic populations. The scatter plots 
depict the joint distributions, which are roughly similar across the real and the synthetic population. Figure 
\ref{scatter_plot_geo} helps visualize the geographic distribution of synthetic households, workplaces and schools in the 
combined districts of Mumbai and Mumbai Suburban. Due to space constraints, the population pyramid and the box plots are placed in the 
\textit{Appendix} (Figure \ref{pop_pyr}, Figure \ref{box_ahw}).

Table \ref{stats} shows results for two statistical tests, the two-sample Kolmogorov–Smirnov test~(KSTest) and the Chi-Squared test~(CSTest). We use the chi-square test for columns with discrete, categorical data (sex in our population). This test returns a $p-$value. A high $p-$value indicates the confidence that the synthetic data and the real data come from the same distribution. We also use the Kolmogorov-Smirnov statistic to compare the marginal distributions of continuous, numerical data (age, height, and weight in our population). We report a score of 1-(KS statistic). A higher score again indicates the confidence that the synthetic data and the real data come from the same continuous distribution.

Another set of tests we conducted to evaluate if our synthetic population can replace the real population to solve a machine 
learning problem. 
We first attempt to predict the weight based on age, sex and height. Similarly, we attempt to predict the height based on age, sex, and weight.
For this experiment, we use two ML models, Linear Regression and Multi-layer Perceptron (MLP) Regressor. First, we train and test the 
models on real data. Next, we train the models on the synthetic data and test the model performance on the real data. The 
comparative results are shown in Table~\ref{ML}.


\subsection{Our Results on BharatSim}\label{bharatsim}
BharatSim, an Agent-based model, is capable of allowing for complex social dynamics. Real-world systems involve interactions between individuals with different attributes (age, weight, etc.) and geographies. These interactions lead to emergent phenomena, while events like a pandemic affect individuals according to their attributes. Agent-based modelling accounts for individual differences and allows us to simulate scenarios of varying complexity. These simulations can guide policy-level interventions (e.g. lockdowns). 

\begin{figure}[htbp]
  \centering
  \includegraphics[width=0.8\linewidth]{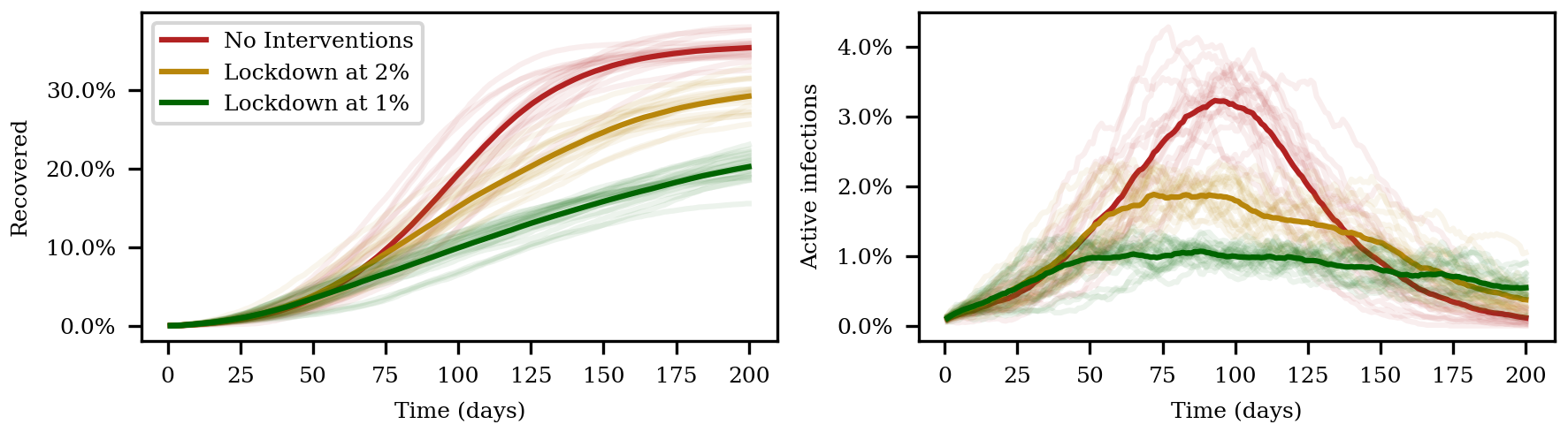}
\caption{No. of recovered individuals (left) and active infections over time}
\label{fig_bharatsim}
\end{figure}

The synthetic population gives us information on the homes and workplaces of all agents in the population, and the individual's movements in a day are charted by defining a ``schedule'' specific to that individual, which decides how individuals move back and forth between their homes and workplaces or schools during a day. The infection spreads stochastically between individuals who share a location at any given time at rates determined by the infection parameters.
In this model, we have constructed a simple SIR compartmental structure~\citet{10.1001/jama.2020.8420} describing a hypothetical disease: Susceptible (S) individuals can be infected at a rate proportional to the fraction of infected (I) individuals they are in contact with. The constant of proportionality, which determines how infectious the disease is, is age-stratified. Once an agent is infected, it can infect other individuals until they recover. The time to recovery is assumed to be drawn from an exponential distribution with a mean of 7 days.
For a simulated lockdown, individual agents' schedules are modified so that they remain at home. All agents are required to follow this rule, except those that are designated by the synthetic population as essential workers and those who have a low probability of adherence to policy-level interventions. Whether or not an individual falls into the latter category is decided by comparing a random coin toss to their "Adherence to interventions" value. The reduction in the number of people in different network locations thus reduces contact between individuals and restricts the spread of the disease.
In our simulations, we have considered two counterfactual scenarios, one in which a lockdown is imposed when the number of active cases is 1\% and 2\% of the total population, respectively. As can be seen in Figure \ref{fig_bharatsim},  the lockdown flattens the curve, keeping the peak in active cases low. Additionally, starting the lockdown earlier is much more effective in curtailing the spread of the disease in our simulation. The results are averaged over 20 runs, with the individual runs in the background. 


\section{Conclusion and Future Work}
In this paper, we propose, SynthPop++, a novel hybrid framework for generating synthetic populations at various scales ranging 
from a few thousand to a billion individuals.  We also suggest a combination of metrics to verify the generated synthetic 
population. 

Future work would include the exploration of methods to generate a synthetic population without the need for sample survey data, as well as the modelling of complex features such as economic variables. Another target would be a realistic intra-city geographical distribution of individuals depending on their demographic and socio-economic attributes. Additionally, some analysis tasks might require a dataset with a specific combination of attributes for which a survey has not been conducted. A method, which can combine two or more datasets with different, partially overlapping sets of attributes, can be used to generate a synthetic population with the specific set of attributes.

\subsubsection*{Acknowledgments}
The authors are grateful for support from the Mphasis F1 Foundation and the Bill and Melinda Gates Foundation, Grant No: R/BMG/PHY/GMN/20.
The authors thank Philip Cherian (Ashoka University) and Gautam I. Menon (Ashoka University and IMSc) for useful discussions. 
The authors also thank the anonymous referees for their useful suggestions. 

\newpage
\bibliography{iclr2023_conference}
\bibliographystyle{iclr2023_conference}

\newpage
\appendix
\section{Appendix}\label{appendix}
\subsection{Synthetic Population Summary Statistics and Charts}
\begin{figure}[H]
\begin{center}
\includegraphics[width=1.0\linewidth]{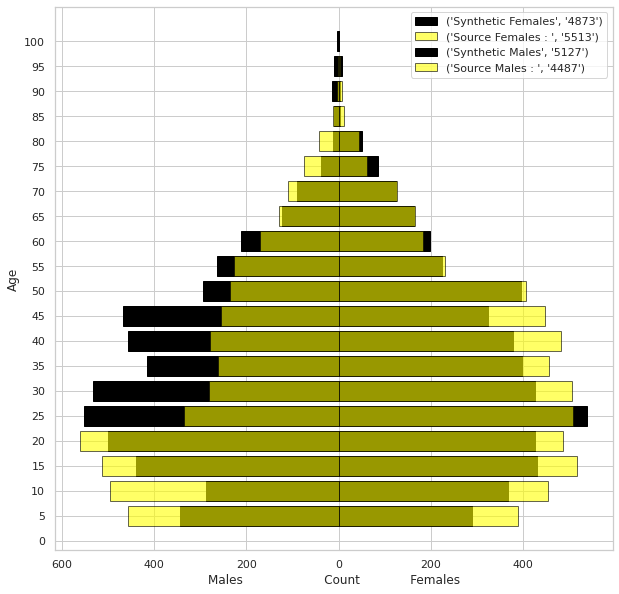} 
\end{center}
\caption{Population pyramid for joint distribution of gender and age: Comparing source population with synthetic population for the combined data for the districts of Mumbai and Mumbai Suburban, India}
\label{pop_pyr}
\end{figure}
\begin{figure}[H]
\begin{center}
\includegraphics[width=1.0\linewidth]{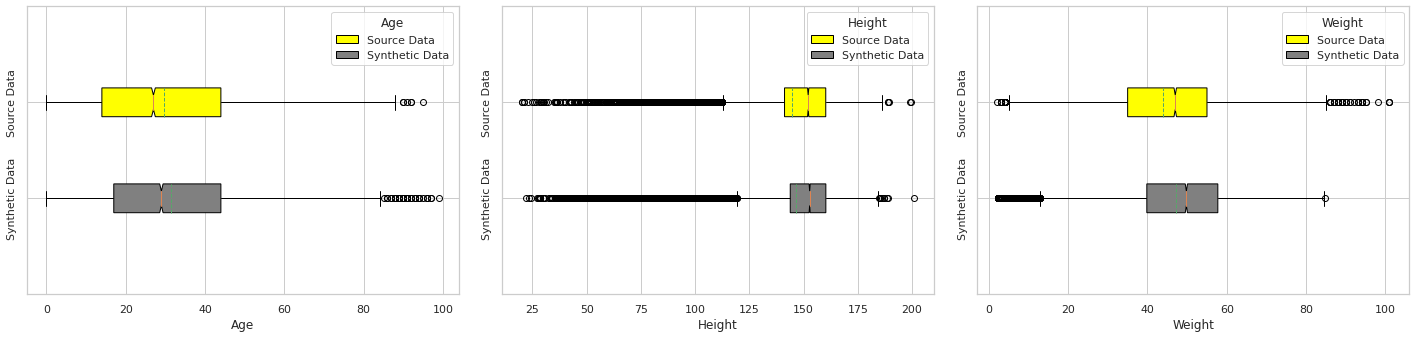} 
\end{center}
\caption{Box plots for marginal distributions of age, height and weight: Comparing source population with synthetic population for the combined data for the districts of Mumbai and Mumbai Suburban, India}
\label{box_ahw}
\end{figure}

\subsection{Sample Population}
A sample of ten individuals in the synthetic population is reproduced below. The columns are split to fit in the page.

\begin{table}[!htb]
		\begin{tabular}{c|c|c|c|c|c|c|c|c}
			Age & SexLabel & Height & Weight & HHID & H\_Lat & H\_Lon & District & AdminUnitName \\
			\hline
			56 & Male & 165.970 & 44.381 & 5.190e+10 & 19.024 & 72.911 & Mumbai & M/E \\
			10 & Male & 138.150 & 34.388 & 5.190e+10 & 19.074 & 72.832 & Mumbai & H/W \\
			3 & Female & 65.460 & 7.205 & 1.038e+11 & 19.258 & 72.867 & Mumbai & R/N \\
			63 & Male & 170.400 & 59.962 & 1.038e+11 & 19.046 & 72.933 & Mumbai & M/E \\
			37 & Female & 143.530 & 60.381 & 5.190e+10 & 19.189 & 72.837 & Mumbai & P/N \\
			46 & Male & 163 & 63.064 & 1.038e+11 & 19.071 & 72.889 & Mumbai & L \\
			35 & Male & 165.660 & 70.111 & 1.038e+11 & 19.073 & 72.906 & Mumbai & N \\
			7 & Female & 119.680 & 21.093 & 1.038e+11 & 19.013 & 72.886 & Mumbai & M/W \\
			29 & Male & 152.520 & 59.198 & 5.190e+10 & 19.131 & 72.898 & Mumbai & S \\
			74 & Female & 169.560 & 37.874 & 5.190e+10 & 18.952 & 72.797 & Mumbai & D \\
		\end{tabular} 
         \end{table}
\begin{table}[!htb]   
		\begin{tabular}{c|c|c|c|c|c|c}
			AdminUnitLatitude & AdminUnitLongitude & Religion & Caste & JobLabel & JobID & WorkPlaceID \\
			\hline
			19.056 & 72.922 & Hindu & other & Carpenters & 81 & 2.001e+12 \\
			19.056 & 72.835 & Hindu & other & Student & 199 & 0 \\
			19.120 & 72.852 & Hindu & other & Homebound & 0 & 0 \\
			19.056 & 72.922 & Hindu & other & Homebound & 0 & 0 \\
			19.188 & 72.842 & Hindu & other & Homebound & 0 & 0 \\
			19.070 & 72.879 & buddhist & other & Labour nec & 99 & 2.001e+12 \\
			19.084 & 72.906 & Hindu & SC & Construction & 95 & 2.001e+12 \\
			19.061 & 72.899 & Hindu & other & Student & 199 & 0 \\
			19.139 & 72.930 & Hindu & other & Construction & 95 & 2.001e+12 \\
			18.963 & 72.813 & Hindu & other & Construction & 95 & 2.001e+12 \\
		\end{tabular} 
    \end{table}
\begin{table}[!htb]        
  
		\begin{tabular}{c|c|c|c|c|c}
			W\_Lat & W\_Lon & essential\_worker & Adherence\_to\_Intervention & PublicTransport\_Jobs & school\_id \\
			\hline
			19.036 & 72.867 & 0 & 0.900 & 1 & 0 \\
			nan & nan & 0 & 0.800 & 1 & 2.001e+12 \\
			nan & nan & 0 & 1 & 1 & 0 \\
			nan & nan & 0 & 1 & 1 & 0 \\
			nan & nan & 0 & 0 & 1 & 0 \\
			19.071 & 72.890 & 0 & 0.900 & 1 & 0 \\
			19.192 & 72.865 & 0 & 0 & 1 & 0 \\
			nan & nan & 0 & 1 & 1 & 2.001e+12 \\
			19.155 & 72.883 & 0 & 0.200 & 1 & 0 \\
			19.024 & 72.850 & 0 & 1 & 1 & 0 \\
		\end{tabular}
  \end{table}
\begin{table}[!htb]
		\begin{tabular}{c|c|c|c|c|c|c}
			school\_lat & school\_long & public\_place\_id & public\_place\_lat & public\_place\_long & Agent\_ID & PSUID \\
			\hline
			nan & nan & 3.001e+12 & 19.024 & 72.910 & 5.191e+10 & 1 \\
			19.177 & 72.867 & 3.001e+12 & 19.179 & 72.828 & 5.191e+10 & 20 \\
			nan & nan & 3.001e+12 & 19.098 & 72.844 & 5.191e+10 & 9 \\
			nan & nan & 3.001e+12 & 19.148 & 72.842 & 5.190e+10 & 14 \\
			nan & nan & 3.001e+12 & 19.189 & 72.830 & 5.190e+10 & 17 \\
			nan & nan & 3.001e+12 & 19.092 & 72.889 & 5.191e+10 & 5 \\
			nan & nan & 3.001e+12 & 19.059 & 72.895 & 5.190e+10 & 8 \\
			18.962 & 72.838 & 3.001e+12 & 19.223 & 72.866 & 5.191e+10 & 22 \\
			nan & nan & 3.001e+12 & 19.129 & 72.913 & 5.191e+10 & 13 \\
			nan & nan & 3.001e+12 & 19.046 & 72.851 & 5.190e+10 & 1 \\
		\end{tabular}
  \end{table}
  \begin{table}[!htb]
	\begin{tabular}{c|c|c|c|c|c|c}
			M\_Fever & M\_Diarrhea & M\_Cataract & M\_Heart\_disease & M\_Diabetes & M\_Leprosy & M\_Cancer \\
			\hline
			0 & 0 & 0 & 0 & 0 & 0 & 0 \\
			0 & 0 & 0 & 0 & 0 & 0 & 0 \\
			0 & 0 & 0 & 0 & 0 & 0 & 0 \\
			0 & 0 & 0 & 0 & 0 & 0 & 0 \\
			0 & 0 & 0 & 0 & 0 & 0 & 0 \\
			0 & 0 & 0 & 0 & 0 & 0 & 0 \\
			0 & 0 & 0 & 0 & 0 & 0 & 0 \\
			0 & 0 & 0 & 0 & 0 & 0 & 0 \\
			0 & 0 & 0 & 0 & 0 & 0 & 0 \\
			0 & 0 & 0 & 0 & 0 & 0 & 0 \\
		\end{tabular}
  \begin{tabular}{c|c|c}
			M\_Asthma & M\_Paralysis & M\_Epilepsy \\
			\hline
			0 & 0 & 0 \\
			0 & 0 & 0 \\
			0 & 0 & 0 \\
			0 & 0 & 0 \\
			0 & 0 & 0 \\
			0 & 0 & 0 \\
			0 & 0 & 1 \\
			0 & 0 & 1 \\
			0 & 0 & 1 \\
			0 & 0 & 0 \\
		\end{tabular}
\end{table}

\end{document}